%% file: root.tex
\documentclass[conference]{IEEEtran}
\IEEEoverridecommandlockouts

\usepackage{cite}
\usepackage{amsmath,amssymb,amsfonts,amsthm}
\usepackage{algorithmic}
\usepackage{graphicx}
\usepackage{textcomp}
\usepackage{xcolor}
\usepackage{hyperref}
\def\BibTeX{{\rm B\kern-.05em{\sc i\kern-.025em b}\kern-.08em
    T\kern-.1667em\lower.7ex\hbox{E}\kern-.125emX}}

\newtheoremstyle{mystyle}
  {}
  {}
  {\itshape}
  {}
  {\bfseries}
  {:}
  { }
  {}
\theoremstyle{mystyle}
\newtheorem{takeaway}{Takeaway}

\graphicspath{{figs/}}
\DeclareGraphicsExtensions{.pdf,.jpeg,.png,.pdf}

\DeclareRobustCommand{\IEEEauthorrefmark}[1]{\smash{\textsuperscript{\footnotesize #1}}}

\input{acronyms}

\begin{document}

\title{Building A Multi-Sensor Platform For Autonomous Driving Research: Challenges and Lessons Learned
\thanks{This research is supported by grants from the Natural Sciences and Engineering Research Council of Canada (NSERC) under grant number \mbox{(RGPIN-2026-05989)} and Canada Research Chair Program \mbox{(CRC-2023-00319)}.}
}

\author{
\IEEEauthorblockN{Paulo~Ricardo~Marques~de~Araujo\IEEEauthorrefmark{1},
Eslam Mounier\IEEEauthorrefmark{1,2},
Qamar Bader\IEEEauthorrefmark{1},\\
Emma Dawson\IEEEauthorrefmark{1},
and Aboelmagd Noureldin\IEEEauthorrefmark{1,3}}
\IEEEauthorblockA{\IEEEauthorrefmark{1} \textit{Queen's University}, Kingston, ON, Canada, \{paulo.araujo, eslam.abdelmoneem, qamar.bader, emma.dawson\}@queensu.ca \\}
\IEEEauthorblockA{\IEEEauthorrefmark{2} \textit{Ain Shams University}, Cairo, Egypt}
\IEEEauthorblockA{\IEEEauthorrefmark{3} \textit{Royal Military College of Canada} , Kingston, ON, Canada, aboelmagd.noureldin@rmc.ca}
}

\maketitle

\begin{abstract}
This paper reports on the challenges encountered and lessons learned during the development and deployment of a flexible multi-sensor platform for autonomous driving research. It aims to serve as a reference for researchers developing new multi-sensor systems. As the need for reliable, diverse datasets increases, novel environments and sensing configurations are essential to tackle real-world operational challenges. Consequently, many research groups create custom multi-sensory data collection platforms, where fundamental issues, such as mechanical design, sensor calibration, power management, and time synchronization, arise regardless of sensor types. We reflect on these challenges and share key insights to guide future platform designs, enhancing reproducibility and robustness in autonomous vehicle testing. We also summarize the mitigation strategies we adopted, for instance, system-wide time synchronization using GNSS timing and NTP protocols, custom calibration routines for different sensor configurations, and design practices to improve reliability and data integrity during field deployments.
\end{abstract}


\section{Introduction}
Autonomous driving has attracted attention both in industry and academia in recent years. This research area involves a diverse set of disciplines, including scene classification, odometry, lane detection, object detection, semantic segmentation, and agent motion prediction~\cite{chen2022milestones}.

The development of autonomous driving algorithms is heavily reliant on high-quality datasets. There is a wide range of datasets available, each focusing on different aspects of autonomous driving. For instance, the KITTI dataset~\cite{geiger2012we} is well known for facilitating research in object detection, tracking, optical flow, depth estimation, and visual odometry, while the Cityscapes dataset~\cite{cordts2016cityscapes} has become a standard benchmark for semantic segmentation due to its high-quality labeled data. Other examples, such as the ApolloScape~\cite{huang2018apolloscape} and the nuScenes~\cite{caesar2020nuscenes} datasets, offer extensive labeling and a large volume of images and multi-modal data, supporting tasks such as 3D map construction, object detection, tracking, and scene understanding. Other examples, such as the Boreas dataset~\cite{burnett2023boreas}, the Hong Kong UrbanNav dataset~\cite{hsu2023hong}, offer multi-modal data collected in multi-season and dense urban environments, facilitating the research of autonomous driving-related tasks under adverse weather conditions and \ac{gnss}-denied environments.


Despite the numerous datasets available, little information has been published on the challenges and best practices involved in this process. For example, the Hong Kong UrbanNav dataset~\cite{hsu2023hong} provides a brief section with valuable information on the lessons they have learned during the creation of the dataset. In order to contribute to this research and address the technical gaps, this work explores and presents the challenges and lessons learned during the development years of the \acs{navinst} experimental setup. The \acs{navinst} dataset~\cite{de2025navinst} has emerged as a new source of data for researchers interested in developing, for example, object detection, odometry, 3D map construction, and indoor positioning algorithms. It features multiple perception and inertial sensors, whose integration and time synchronization were sources of many challenges, as discussed in the following sections. Our objective with this paper is to share the insights gained in the development of the system in order to help practitioners interested in developing their own setup rigs to contribute to the autonomous driving research area. 

\begin{figure*}
    \centering
    \includegraphics[width=0.75\linewidth]{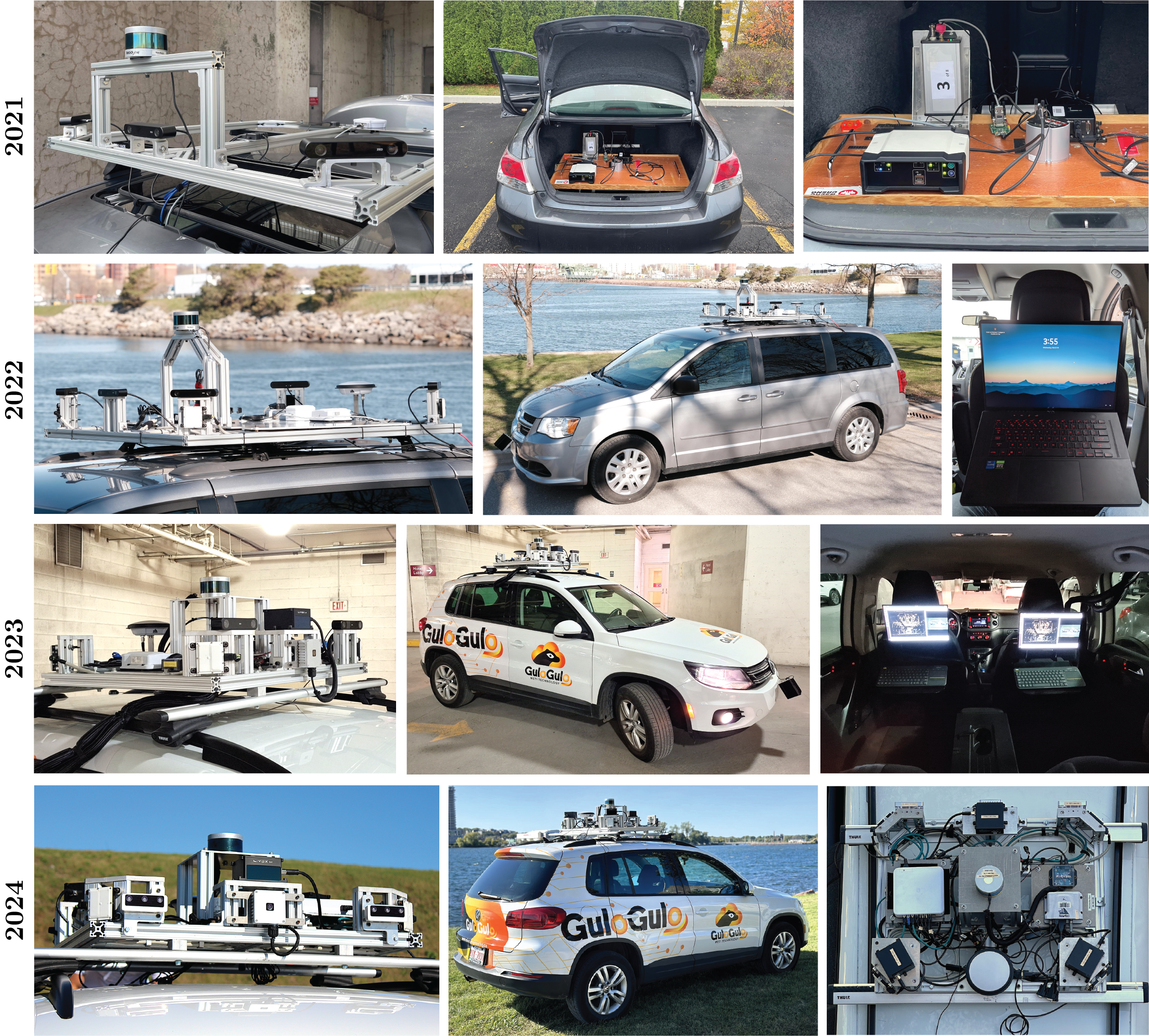}
    \caption{Evolution of the NavINST experimental setup over time.}
    \label{fig:setups}
\end{figure*}

\section{Development Timeline}

The \acf{navinst} laboratory located at the Royal Military College of Canada is well known for its work on \ac{gnss}, inertial navigation, multi-sensor systems, and related areas. 
To expand the research possibilities, a project of a multi-sensor experimental setup was defined.
The development of the experimental setup was driven by some initial pillars: (1) it must be portable and flexible, and (2) it must include perception sensors such as cameras, \ac{lidar}, and \ac{radar}. The first pillar was established, taking into account that the team would need to travel to different cities in Canada to collect data for different research purposes. The second pillar served as the backbone for the initial development of the system, which was mainly inspired by existing datasets in mid-2020. In addition, the number of datasets that offered \ac{radar} data collected in urban environments at that time was limited, and this aspect became an additional motivation.

The experimental setup has evolved in recent years, as illustrated in \figurename~\ref{fig:setups}. The initial version, 2021, was inspired by previous experimental setups available in the laboratory for \ac{gnss}-based research. The reference system, composed of a high-end \ac{gnss} receiver and a high-end \ac{imu} was located in the trunk close to the rear axle of the vehicle. Three stereo cameras, one mechanical \ac{lidar}, and a \ac{gnss} antenna were installed on the roof.

In the second version, 2022, more stereo cameras along with \acp{radar} and other \acp{imu} were installed. All sensors were moved to the roof to comply with the first development pillar, as detailed in the next section. In 2023, the third version was implemented with redesigned and optimized computing systems, updated stereo cameras, and the addition of a monocular camera and a modern solid-state \ac{lidar}. The latest version of the experimental setup was deployed in 2024 and featured three solid-state \acp{lidar} and new stereo cameras.



\section{Challenges and Lessons Learned}


\subsection{Mechanical Design}
The plan was to travel with the setup safely to different places, which would be achieved if the main components were stored inside the vehicle. Therefore, the testbed was designed using aluminum frames due to their ease-of-use capabilities and flexibility in accommodating new components. The main disadvantage, however, was the total weight of the system, which in the 2022 version reached approximately $40$\,kg with all installed equipment.

The main change in mechanical design happened between the 2021 and 2022 versions. The early version had part of the sensors installed in the trunk, with several sensors mounted on a wooden board. This configuration was convenient to measure vehicle dynamics using the \acp{imu}, but it was significantly complicated to calibrate the board to avoid misalignment and erroneous dynamics measurements due to the gravity field. Therefore, from 2022 on, all sensors were mounted together on a single test bed.

\begin{takeaway}
    For sensor placement, it is important to take into consideration aspects such as installation accessibility and calibration of extrinsic transformations.
\end{takeaway}

\begin{figure}
    \centering
    \includegraphics[width=0.7\columnwidth]{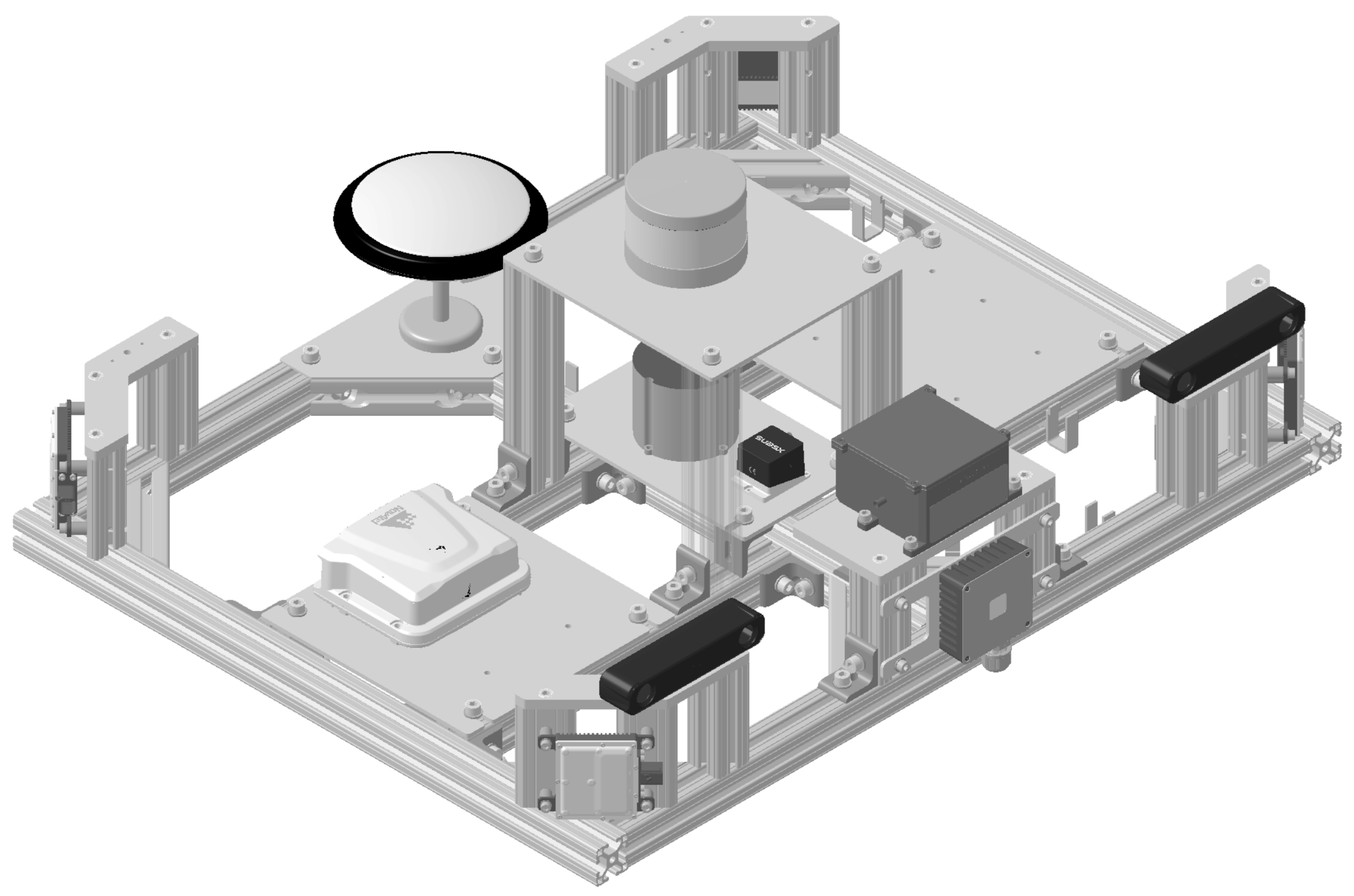}
    \caption{Digital twin of the 2023 experimental setup.}
    \label{fig:digital-twin}
\end{figure}

A key component of mechanical design was the use of \ac{cad} systems to build digital twins of testbeds, as illustrated in \figurename~\ref{fig:digital-twin}. The digital twin allowed for the detection of problems related to sensor field-of-view blockage, and enabled the design of mechanical parts to be more accurate, saving time and money.

\begin{takeaway}
    Knowledge of mechanical and \ac{cad} systems can significantly accelerate the development process, allowing valuable resources to be allocated to other critical components of the system.
\end{takeaway}

Since 2022, the experimental setup has traveled to different cities in Canada. Due to its flexibility, it was possible to pack and ship it by airplane to a different city, where the local team assembled it following predefined instructions and collected data successfully. As can be seen in \figurename~\ref{fig:setups}, different vehicles were used to collect data, highlighting the flexibility of having a unified system installed on the roof of these vehicles.
It is important to note that each time the experimental setup, including frames and sensors, is disassembled and reassembled, the extrinsic transformations of the system must be re-calibrated.

\subsection{Calibration}
The calibration process of the experimental setup also evolved with time. The 2021 version relied on rough measurements of the cameras and \ac{lidar} distances relative to the system reference point. In 2022, the addition of other sensors, especially \acp{radar}, demanded more sophisticated approaches. That system was calibrated using an open-source toolbox capable of computing the extrinsic transformations between a mechanical \ac{lidar}, a \ac{radar}, and a stereo camera~\cite{domhof2019extrinsic}. The extrinsic transformations of each camera and \ac{radar} pair were computed individually with the mechanical \ac{lidar}. The main disadvantage of this approach is the propagation of uncertainties in the transformation tree.
For example, transforming points between the coordinate systems of \ac{radar} $a$ \ac{radar} $b$ involves the compounded uncertainties of the transformations from each \ac{radar} to the \ac{lidar} frame.

\begin{takeaway}
    Uncertainty propagation analysis can improve the identification of global uncertainties, improving the interpretation of future results. For example, in a \ac{lidar}-camera positioning system, the desired accuracy may be unattainable if the uncertainty of the extrinsic transformation already exceeds the desired threshold.
\end{takeaway}

\begin{figure}
    \centering
    \includegraphics[width=0.75\columnwidth]{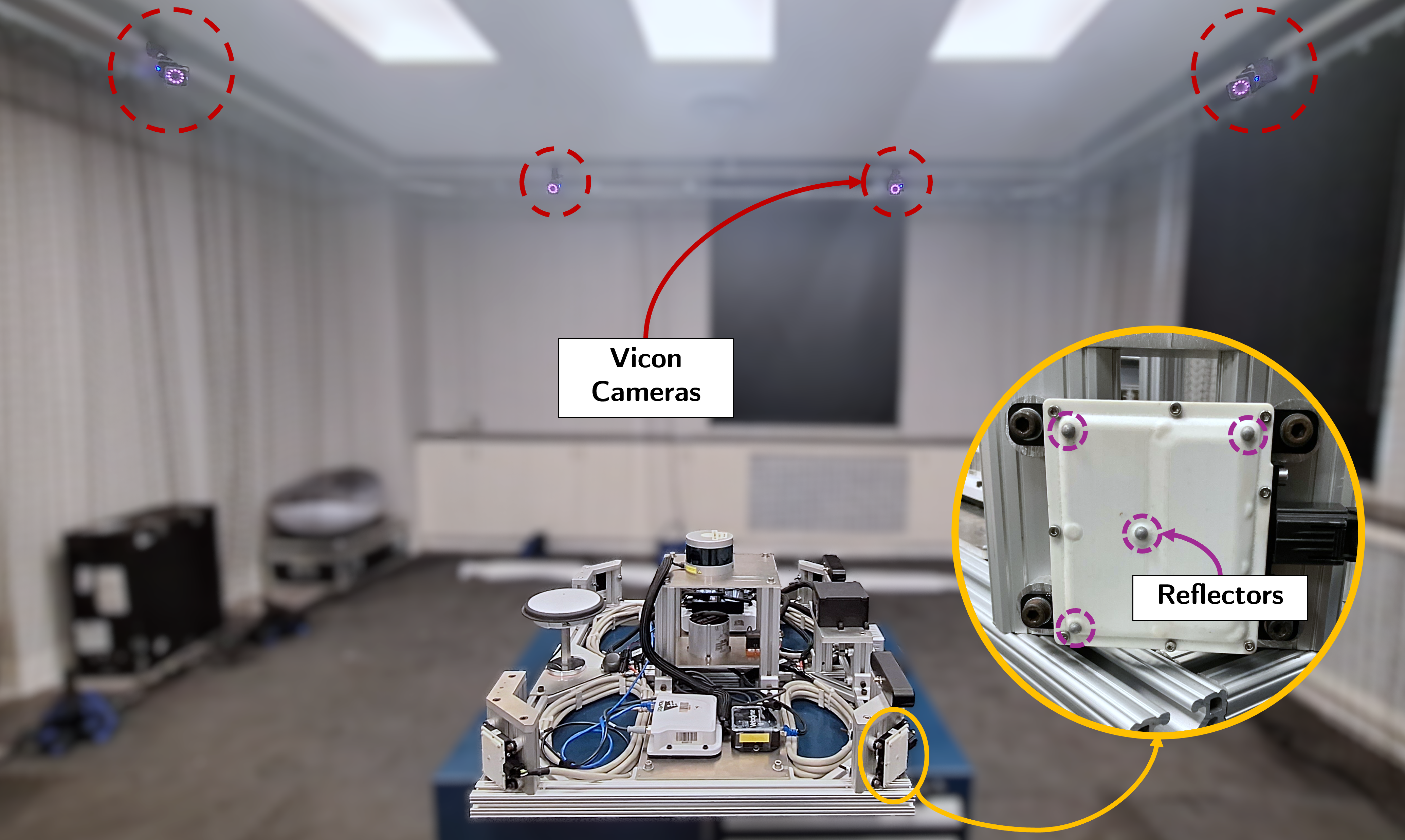}
    \caption{Calibration of some extrinsic transformations using a camera tracking system by Vicon. Reflective markers were mounted on the sensors to enable precise calibration.}
    \label{fig:vicon}
\end{figure}

In 2023, the stereo cameras did not share the same field of view as the \acp{radar}, necessitating a different approach. The setup was redesigned and carefully assembled to match the dimensions of its digital twin. Main dimensions were obtained via laser measurements, while more complex transformations were verified using the Vicon tracking system, as presented in \figurename~\ref{fig:vicon}.
The Vicon system offers sub-millimeter accuracy, facilitating extrinsic calibration through the use of reflective spheres. While not strictly mandatory, it is recommended to utilize such a system if available. Since the setup was designed with flexibility and transportability in mind, Vicon-based verification could be performed indoors prior to installation on the vehicle. Finally, the intrinsic calibration of the cameras and the extrinsic calibration between the cameras and the mechanical \ac{lidar} were performed using MATLAB toolboxes.

Finally, the solid-state \ac{lidar} posed new challenges. Existing toolboxes designed for mechanical \acp{lidar} failed to deliver reliable results when tested with solid-state \acp{lidar}. Therefore, a custom calibration algorithm was developed to calibrate each solid-state \ac{lidar} with the mechanical \ac{lidar}. After this process, the transformation tree was complete and ready to be used.

\begin{takeaway}
    New sensors may require the development of new calibration algorithms, which may consume more time than initially predicted, depending on the complexity of the system.
\end{takeaway}

\begin{takeaway}
    More sensors, more challenges, and difficulties in calibrating the experimental setup. The relationship between time needed and the number of sensors seems to be linear, but it is not.
\end{takeaway}


\subsection{Reference System}

Similarly to other works, the reference of the \ac{navinst} dataset is based on a \ac{gnss}-\ac{imu}-based system that includes a high-end receiver and a tactical-grade \ac{imu}. These systems are robust and may not pose difficulties under normal operating conditions.

The main challenge with our system was interference caused by other sensors. As explained in Section~\ref{sec:other-considerations}, the first cameras employed in the setup used \ac{usb} 3.0 for data transfer.
However, \ac{usb}~3.0 can generate radio frequency noise \cite{sun2019mechanism}. In our case, the high data rate of the cameras combined with \ac{usb}~3.0 operation was found to introduce \ac{emi} that adversely affected the \ac{gnss} antenna.

\begin{figure}
    \centering
    \includegraphics[width=\linewidth]{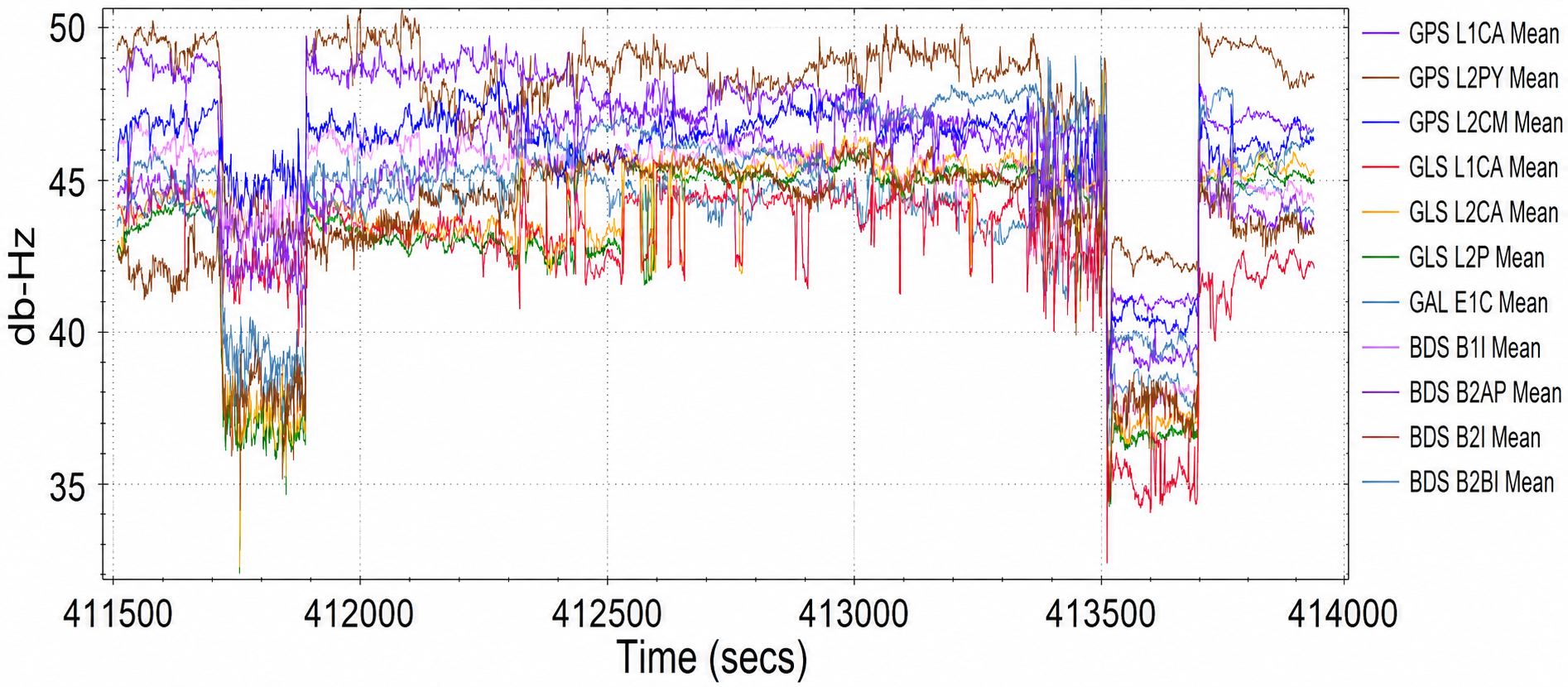}
    \caption{Carrier-to-noise ratio under interferences.}
    \label{fig:carrier-to-noise}
\end{figure}

This issue was identified after months of investigation and iteration with the manufacturer. The interference mainly affected satellites below $30^\circ$ elevation, hindering the system from initializing and converging properly under certain circumstances. In addition, the mechanical \ac{lidar} was also a source of \ac{emi} due to its limited shielding. Newer and more advanced \acp{lidar} systems are expected to offer improved robustness with reduced impact on onboard \ac{gnss} systems.  \figurename~\ref{fig:carrier-to-noise} illustrates the effect of interference on the carrier-to-noise ratio. The periods when the mechanical \ac{lidar} and the cameras were active can be clearly distinguished. Specifically, we observed a degradation of approximately $10$ dB-Hz in the carrier-to-noise density ratio (C/N0) for the affected signals, dropping from a nominal range of $45$–$50$ dB-Hz to $35$–$40$ dB-Hz during sensor operation.

\begin{takeaway}
    If possible, avoid \ac{usb} 3.0 devices, as the volume of data traveling at high speed will certainly cause \ac{emi} issues. If not possible, mitigate interference by shielding the devices with copper tape, using shielded cables, and adding ferrite beads to the cables— especially close to the terminals. Additionally, try to keep the \ac{gnss} antenna as far as possible from these devices.
\end{takeaway}

\ac{gnss}-based measurements are affected by ionospheric activity. This aspect can vary depending on the geographic location of the experimental setup. For example, Novatel offers a tool\footnote{\href{https://novatel.com/tech-talk/ionospheric-scintillation/ionospheric-activity-forecast-tool}{Ionospheric Activity Forecast Map: https://novatel.com/tech-talk/ionospheric-scintillation/ionospheric-activity-forecast-tool}} that can help plan data collection campaigns to avoid periods of high ionospheric activity.
In addition to this verification, receivers feature other functionalities that monitor the accuracy of signals, correct for interferences, and improve integrated solutions.

\begin{takeaway}
    For \ac{gnss}-\ac{imu}-based reference systems, check the user manual to determine which calibration routines are available. These may include the estimation of the lever arm between the receiver and the antenna, as well as the rotational offset between the vehicle and the \ac{imu} frame. The accuracy of the results can improve substantially.
\end{takeaway}

As mentioned earlier, the setup was deployed in many complex environments and it became evident that the reference system struggled in dense areas occupied by high-rise buildings when operating in standalone mode. To support the reference system in such conditions, we employed stationary base stations. These enabled the post-processing of trajectories to correct for different \ac{gnss}-related problems caused by limited sky visibility. In our dataset, the proprietary Novatel Inertial Explorer software was used, but other solutions such as the open-source RTKLib\footnote{\href{https://www.rtklib.com/}{https://www.rtklib.com/}} can be an easy-to-use and affordable alternative. In some cases, during data collection, there was a team on the ground taking care of a temporary base station installed in a strategic location close to the trajectory, while another team was driving the vehicle and monitoring the systems.

\begin{takeaway}
    The performance of the post-processed corrections is maximized when the base station is close to the vehicle; keeping a maximum distance of $10$~km is a good rule of thumb.
\end{takeaway}

\subsection{System Architecture}

\subsubsection{Hardware-Related}
Managing a complex multi-sensor system often requires distributing the load across multiple computing units, each of which is responsible for a subset of sensors. However, this can introduce challenges related to hardware integration, mounting constraints, and sensor data synchronization.

In 2021, our initial system relied on multiple laptops. This approach was convenient in the early stages, providing the necessary \ac{hmi} capabilities to handle data acquisition for specific sensors on each laptop. Basic time synchronization was facilitated using an external \ac{gnss} receiver connected to each laptop, allowing \ac{gnss} time to be locally logged.

As the system evolved and new sensors were integrated, the need for more computational power and better communication emerged. In 2022, we continued using laptops and we created a unified Ethernet network to enable inter-device connectivity and facilitate synchronization between subsystems.

In 2023, 
a main computer handled most sensor acquisition, while an NVIDIA Orin platform was dedicated to processing data from demanding stereo cameras. This transition simplified hardware placement in the trunk of the vehicle and the \ac{hmi}.
Importantly, stereo cameras were connected via \ac{gmsl}, eliminating \ac{usb} 3.0 devices and thus resolving the aforementioned \ac{emi} issues. Most of the remaining sensors shared the unified Ethernet network, while devices with a lower bandwidth, such as \ac{gnss} receivers and odometers, utilized \ac{usb} 2.0.

\begin{figure}
    \centering
    \includegraphics[trim={0.5cm 0.5cm 0.4cm 0.4cm}, clip, width=0.9\columnwidth]{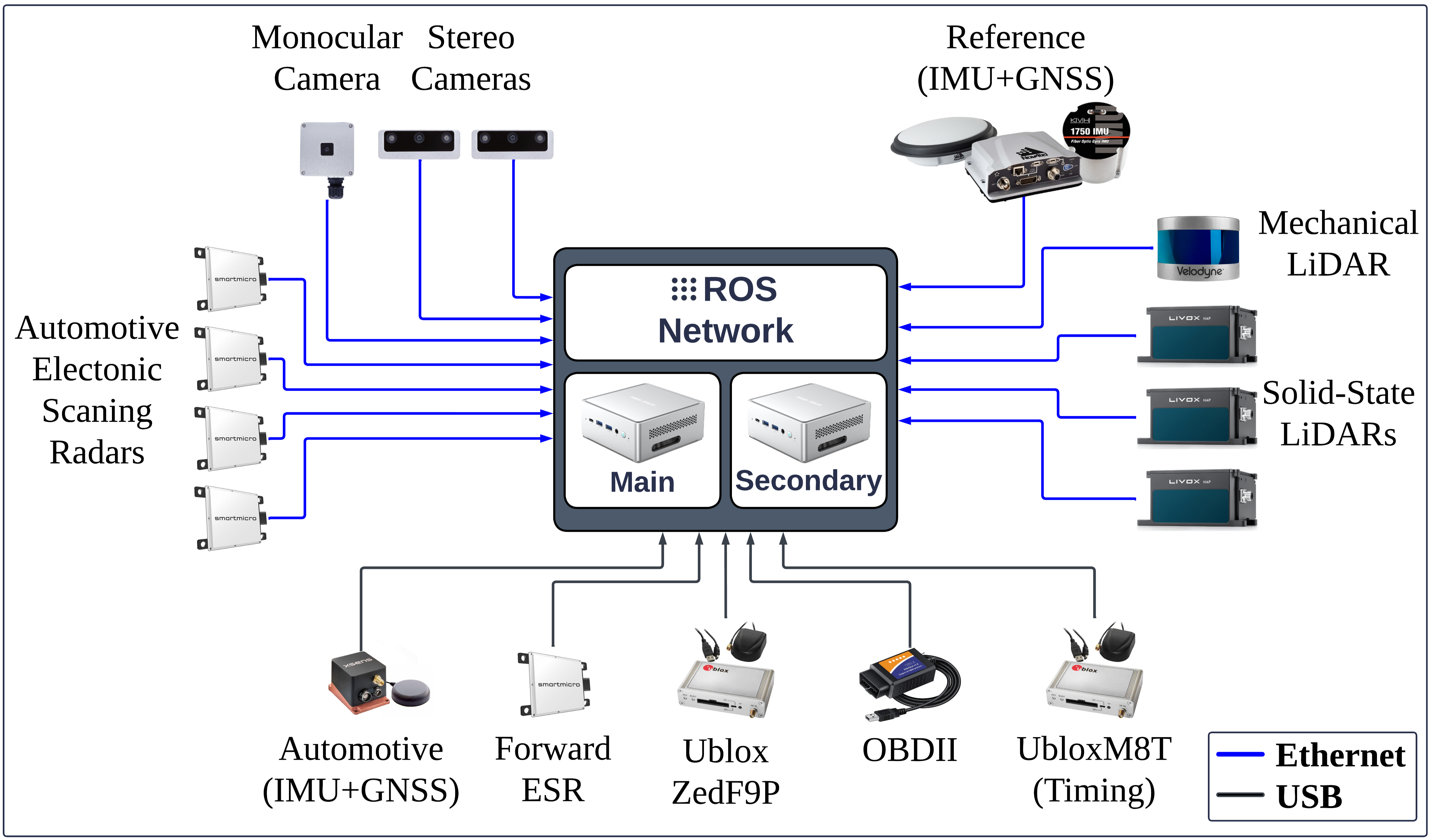}
    \caption{Current system architecture.}
    \label{fig:system-architecture}
\end{figure}

In the last update in 2024, we acquired new stereo cameras with onboard processing and \ac{poe} support, eliminating the need for \ac{gmsl} and dedicated computing units. The updated architecture, shown in \figurename~\ref{fig:system-architecture}, was consolidated into the unified Ethernet network, connecting two compact high-performance computers and advanced sensors, including monocular and stereo cameras based on \ac{poe}, and multiple solid-state \acp{lidar}. Some low-bandwidth sensors remain connected through \ac{usb} 2.0 directly to computers.

\begin{takeaway}
    The continuous and reliable operation of a multi-sensor system depends on an effective architecture, which involves, but is not limited to, computational load distribution, hardware integration complexity, and \ac{hmi} practicality.
\end{takeaway}

\subsubsection{Software-Related}
Maintaining a unified time reference throughout the multi-sensor system is a critical requirement to ensure precise synchronization during data acquisition. One lesson learned from our early 2021 setup is the limitation of relying on proprietary software running independently on each computer for reliable time synchronization.

A major improvement introduced in 2022 was the adoption of \ac{ros}, a widely used middleware in robotics and autonomous systems \cite{quigley2009ros}. From 2022 to 2024, \ac{ros} served as the backbone of our software architecture, supporting the increasing complexity of the system. Its modular design allowed for the configuration of dedicated software drivers for each sensor, effectively abstracting the underlying hardware complexity. This facilitated seamless integration and communication between sensors and computing units throughout the system.

Due to the widespread adoption of \ac{ros}, most sensor vendors offered \ac{ros}-compatible drivers for configuration, control, and data acquisition. However, in several cases, the provided drivers required source code modifications, or entirely new drivers had to be developed.

\begin{takeaway}
    The unmentioned sensors did not pose complex challenges and their learning curves were reasonable. However, it is strongly recommended to verify whether the manufacturers provide \ac{ros} drivers before purchasing the sensors, especially if the team plans to use \ac{ros} as the backbone system.
\end{takeaway}

Although \ac{ros} facilitates message synchronization between sensors and their directly connected host computers, it does not inherently ensure consistent timestamps in a distributed system. Consequently, clock drift between multiple computers can still occur, potentially degrading synchronization accuracy, particularly over long-duration experiments. To mitigate this problem, system-wide time synchronization must be implemented. For this purpose, we used well-established Linux tools, including \textit{Chrony} and \textit{GPSd} \cite{martins2024time}. In our setup, a computer was designated as the time master, running the \textit{GPSd} service connected to a dedicated \ac{gnss} receiver to obtain precise timing information. The \textit{Chrony} service was then used to align this computer clock with \ac{gnss} time and configure it as a local \ac{ntp} server. Other computers in the system were configured as \ac{ntp} clients, synchronizing their clocks with the local \ac{ntp} server to maintain consistent time throughout the system. In a Local Area Network (LAN) scenario, NTP can maintain 1-millisecond accuracy~\cite{martins2024time}. This configuration allowed for data collection using the \ac{ros} timeline with minimal lag.

\subsection{Powering the System}
Providing power to the computers and sensors on board is another important task that requires planning. In the early versions of the experimental setup, the lab did not have a fixed vehicle. Therefore, to avoid modifications to the vehicle, the sensors and computers were powered using portable batteries. The inconvenient part was to carry multiple batteries inside the vehicle for long data collection campaigns, especially when traveling to distant cities. In addition, batteries would always die in the middle of a trajectory, ruining the entire run.

As mentioned in Section~\ref{sec:other-considerations}, laptops collecting camera data were power hungry and significantly impacted battery life during experiments. A solution was to use a charger connected to the DC power of the vehicle to slow down their discharge rate.

In 2023, a dedicated vehicle arrived at the lab and modifications were allowed. In the trunk, a power inverter connected to the vehicle's alternator was installed, significantly simplifying power distribution and cable management.
The only downside of this approach is that the engine must be running at all times, even during debug and testing periods.


\begin{takeaway}
    Relying on the vehicle's power is an interesting and reliable approach. To overcome the inconvenience of keeping the engine running during tests, especially in confined areas such as garages in the winter, a power adapter can be installed to enable the vehicle's battery to be charged using power connectors on the wall.
\end{takeaway}

\subsection{Other Considerations} \label{sec:other-considerations}

The complexity of the system dictates the volume of tasks needed for data collection. Therefore, it is important to test and document the steps that every team member must follow during data acquisition. This step is of utmost importance, especially in research laboratories, where researchers stay for a short period, but it is important to ensure that the knowledge they develop is not lost.

\begin{takeaway}
    A clear and easy-to-use checklist can save time during data acquisition and prevent errors that can compromise the entire campaign, e.g., initializing the recording without certifying that all sensors are visible in the network.
\end{takeaway}

Metadata was shown to be an important asset for the verification of the collected data. It is important to verify the integrity of the data as soon as a trajectory is concluded. However, in practice, time constraints may delay this process. Therefore, maintaining a logbook with detailed information about each run can help others during post-processing or using the data. For example, in some situations, the computer used to record the cameras crashed in the middle of a trajectory. Without a logbook explaining what happened on that trajectory, it would be nearly impossible for other researchers to identify or understand a gap in a sensor timeline.

\begin{takeaway}
    A logbook can be a simple text document that contains information about the researchers present in the experiment, the sensors used, the time of day, the weather conditions, and the quality of the reference solution. However, it can be a more sophisticated system using libraries such as Docker to automate the information collection~\cite{vizzo2023toward}.
\end{takeaway}

Finally, with regard to sensors and hardware,
in 2021, our initial camera rig consisted of three StereoLabs ZED 2i stereo cameras.
These units were mounted on the vehicle roof to maximize the forward field of view, as shown in \figurename~\ref{fig:setups}. In 2022, two additional ZED 2i cameras were integrated into the system, where
each camera communicated over \ac{usb}‑C 3.1 Gen 1 (5 Gbps), resulting in an aggregate raw data rate of approximately 25 Gbps ($\sim$3.1 GB/s) before compression, imposing significant demands on our SSD storage and real‑time data pipeline.
To manage this throughput, which averaged approximately 3 GB/s, we utilized high-speed NVMe SSDs to prevent I/O bottlenecks. This hardware choice was critical to ensure that the write speed exceeded the data generation rate, preventing buffer overflows and frame drops.

\begin{takeaway}
    The deployment of multiple high-resolution stereo cameras can quickly saturate \ac{usb} and storage bandwidth. Considering high-throughput interfaces such as PCIe, GSML, and 10 GbE is important for system scalability.
\end{takeaway}

In 2023, we transitioned to the ZED X stereo camera, which employs GMSL2 coaxial cables capable of 6 Gbps per link with robust shielding and supports hardware-triggered frame synchronization via a dedicated sync header. An NVIDIA Orin offloaded stereo depth computation and video encoding, reducing CPU load on the primary compute module.

\begin{takeaway}
    Even with hardware-trigger support, multi-camera rigs can exhibit timing bias due to differences in exposure start, image pipeline buffering, and frame-grab intervals. Characterize per-camera latency and apply software compensation when needed.
\end{takeaway}

In early 2024, we adopted Luxonis OAK‑D stereo cameras, which used Ethernet for data acquisition. These cameras facilitated the system architecture as presented in \figurename~\ref{fig:system-architecture}.

\begin{takeaway}
    Before investing in the purchase of a sensor, it is important to verify the support available by the manufacturer and its integration impact in the setup.
\end{takeaway}


\section{Conclusion}
This paper presented a report on the development and deployment of a modular, multi-sensor platform designed to support autonomous driving research in GNSS-challenged environments.
Key challenges addressed during the development phase included mechanical integration, sensor calibration, \ac{emi} mitigation, system time synchronization, and power distribution. 
Solutions such as the use of digital twins for mechanical design, custom calibration routines for unconventional sensor configurations, and system-wide time synchronization using GNSS timing and NTP protocols were all implemented to improve accuracy and reliability.
The lessons learned throughout this process are intended to guide future efforts in multi-sensor system design, with an emphasis on scalability, robustness, and data integrity in real-world driving conditions.

\bibliographystyle{IEEEtran}
\bibliography{refs}

\end{document}

%% file: acronyms.tex
\usepackage{acro}
\DeclareAcronym{navinst}{
  short = NavINST,
  long  = Navigation and Instrumentation}
\DeclareAcronym{gnss}{
  short = GNSS,
  long  = global navigation satellite system}
  \DeclareAcronym{lidar}{
  short = LiDAR,
  long  = light detection and ranging}
  \DeclareAcronym{radar}{
  short = RADAR,
  long  = radio detection and ranging}
\DeclareAcronym{av}{
  short = AV,
  long  = autonomous vehicle}
\DeclareAcronym{imu}{
  short = IMU,
  short-plural-form = IMUs,
  long  = inertial measurement unit,
  long-plural-form = inertial measurement units}
\DeclareAcronym{ros}{
  short = ROS,
  long  = robot operating system}
\DeclareAcronym{esr}{
  short = ESR,
  short-plural-form = ESRs,
  long  = electronically scanning RADAR,
  long-plural-form = electronically scanning RADARs}
\DeclareAcronym{usb}{
  short = USB,
  long = universal serial bus
}
\DeclareAcronym{gmsl}{
  short = GMSL,
  long = gigabit multimedia serial link
}
\DeclareAcronym{ntp}{
  short = NTP,
  long = network time protocol
}
\DeclareAcronym{cad}{
  short = CAD,
  long = computer-aided design
}
\DeclareAcronym{emi}{
  short = EMI,
  long = electromagnetic interference
}

\DeclareAcronym{hmi}{
  short = HMI,
  long = human-machine interface
}

\DeclareAcronym{poe}{
  short = PoE,
  long = power-over-Ethernet
}


%% file: refs.bib
@article{chen2022milestones,
  title={Milestones in autonomous driving and intelligent vehicles: Survey of surveys},
  author={Chen, Long and Li, Yuchen and Huang, Chao and Li, Bai and Xing, Yang and Tian, Daxin and Li, Li and Hu, Zhongxu and Na, Xiaoxiang and Li, Zixuan and others},
  journal={IEEE Transactions on Intelligent Vehicles},
  volume={8},
  number={2},
  pages={1046--1056},
  year={2022},
  publisher={IEEE}
}

@inproceedings{geiger2012we,
  title={Are we ready for autonomous driving? the kitti vision benchmark suite},
  author={Geiger, Andreas and Lenz, Philip and Urtasun, Raquel},
  booktitle={2012 IEEE conference on computer vision and pattern recognition},
  pages={3354--3361},
  year={2012},
  organization={IEEE}
}

@inproceedings{cordts2016cityscapes,
  title={The cityscapes dataset for semantic urban scene understanding},
  author={Cordts, Marius and Omran, Mohamed and Ramos, Sebastian and Rehfeld, Timo and Enzweiler, Markus and Benenson, Rodrigo and Franke, Uwe and Roth, Stefan and Schiele, Bernt},
  booktitle={Proceedings of the IEEE conference on computer vision and pattern recognition},
  pages={3213--3223},
  year={2016}
}

@inproceedings{caesar2020nuscenes,
  title={nuscenes: A multimodal dataset for autonomous driving},
  author={Caesar, Holger and Bankiti, Varun and Lang, Alex H and Vora, Sourabh and Liong, Venice Erin and Xu, Qiang and Krishnan, Anush and Pan, Yu and Baldan, Giancarlo and Beijbom, Oscar},
  booktitle={Proceedings of the IEEE/CVF conference on computer vision and pattern recognition},
  pages={11621--11631},
  year={2020}
}

@inproceedings{huang2018apolloscape,
  title={The apolloscape dataset for autonomous driving},
  author={Huang, Xinyu and Cheng, Xinjing and Geng, Qichuan and Cao, Binbin and Zhou, Dingfu and Wang, Peng and Lin, Yuanqing and Yang, Ruigang},
  booktitle={Proceedings of the IEEE conference on computer vision and pattern recognition workshops},
  pages={954--960},
  year={2018}
}

@article{burnett2023boreas,
  title={Boreas: A multi-season autonomous driving dataset},
  author={Burnett, Keenan and Yoon, David J and Wu, Yuchen and Li, Andrew Z and Zhang, Haowei and Lu, Shichen and Qian, Jingxing and Tseng, Wei-Kang and Lambert, Andrew and Leung, Keith YK and others},
  journal={The International Journal of Robotics Research},
  volume={42},
  number={1-2},
  pages={33--42},
  year={2023},
  publisher={SAGE Publications Sage UK: London, England}
}

@article{de2025navinst,
  title={The NavINST dataset for multi-sensor autonomous navigation},
  author={De Araujo, Paulo Ricardo Marques and Mounier, Eslam and Bader, Qamar and Dawson, Emma and Abdelaziz, Shaza I Kaoud and Zekry, Ahmed and Elhabiby, Mohamed and Noureldin, Aboelmagd},
  journal={IEEE Access},
  year={2025},
  publisher={IEEE}
}

@inproceedings{domhof2019extrinsic,
  title={An extrinsic calibration tool for radar, camera and lidar},
  author={Domhof, Joris and Kooij, Julian FP and Gavrila, Dariu M},
  booktitle={2019 International Conference on Robotics and Automation (ICRA)},
  pages={8107--8113},
  year={2019},
  organization={IEEE}
}

@inproceedings{vizzo2023toward,
  title={Toward Reproducible Version-Controlled Perception Platforms: Embracing Simplicity in Autonomous Vehicle Dataset Acquisition},
  author={Vizzo, Ignacio and Mersch, Benedikt and Nunes, Lucas and Wiesmann, Louis and Guadagnino, Tiziano and Stachniss, Cyrill},
  booktitle={2023 IEEE 26th International Conference on Intelligent Transportation Systems (ITSC)},
  pages={5838--5845},
  year={2023},
  organization={IEEE}
}

@inproceedings{quigley2009ros,
  title={{ROS}: an open-source Robot Operating System},
  author={Quigley, Morgan and Conley, Ken and Gerkey, Brian and Faust, Josh and Foote, Tully and Leibs, Jeremy and Wheeler, Rob and Ng, Andrew Y and others},
  booktitle={ICRA workshop on open source software},
  volume={3},
  number={3.2},
  pages={5},
  year={2009},
  organization={Kobe, Japan}
}

@article{hsu2023hong,
  title={Hong Kong UrbanNav: An open-source multisensory dataset for benchmarking urban navigation algorithms},
  author={Hsu, Li-Ta and Huang, Feng and Ng, Hoi-Fung and Zhang, Guohao and Zhong, Yihan and Bai, Xiwei and Wen, Weisong},
  journal={NAVIGATION: Journal of the Institute of Navigation},
  volume={70},
  number={4},
  year={2023},
  publisher={Institute of Navigation}
}

@article{sun2019mechanism,
  title={Mechanism and validation of USB 3.0 connector caused radio frequency interference},
  author={Sun, Yin and Lin, Hank and Tseng, Bin-Chyi and Pommerenke, David and Hwang, Chulsoon},
  journal={IEEE Transactions on Electromagnetic Compatibility},
  volume={62},
  number={4},
  pages={1169--1178},
  year={2019},
  publisher={IEEE}
}

@inproceedings{martins2024time,
  title={Time synchronization in V2X communications},
  author={Martins, Rodrigo and Raposo, Duarte and Lopes, Rui and Rito, Pedro and Sargento, Susana},
  booktitle={2024 IEEE Vehicular Networking Conference (VNC)},
  pages={25--32},
  year={2024},
  organization={IEEE}
}
